\documentclass{article}

\usepackage{PRIMEarxiv}

\usepackage[utf8]{inputenc} 
\usepackage[T1]{fontenc}    
\usepackage{hyperref}       
\usepackage{url}            
\usepackage{booktabs}       
\usepackage{amsfonts}       
\usepackage{nicefrac}       
\usepackage{microtype}      
\usepackage{lipsum}
\usepackage{fancyhdr}       
\usepackage{graphicx}       
\usepackage[table,xcdraw]{xcolor}
\usepackage{amsmath}
\usepackage{booktabs} 

\usepackage{algorithm}
\usepackage{algpseudocode}
\usepackage{listings}
\usepackage{caption}
\usepackage{geometry}
\usepackage{siunitx} 
\usepackage{multirow}

\title{Zero-Shot Decision Tree Construction\\ via Large Language Models
}

\author{
  Lucas Carrasco, 
  Felipe Urrutia, 
  Andrés Abeliuk\thanks{Corresponding author:\textit{ aabeliuk@dcc.uchile.cl}}\\
  Department of Computer Science, University of Chile\\ 
  National Center for Artificial Intelligence (CENIA) \\
  Santiago, Chile\\
}

\begin{document}

\maketitle

\begin{abstract}
 This paper introduces a novel algorithm for constructing decision trees using large language models (LLMs) in a zero-shot manner based on Classification and Regression Trees (CART) principles. Traditional decision tree induction methods rely heavily on labeled data to recursively partition data using criteria such as information gain or the Gini index. In contrast, we propose a method that uses the pre-trained knowledge embedded in LLMs to build decision trees without requiring any training data. Our approach leverages LLMs to perform operations essential for decision tree construction, including attribute discretization, probability calculation, and Gini index computation based on the probabilities. We show that these zero-shot decision trees can outperform baseline zero-shot methods and achieve competitive performance compared to supervised data-driven decision trees on tabular datasets. The decision trees constructed via this method provide transparent and interpretable models, addressing data scarcity while preserving interpretability. This work establishes a new baseline in low-data machine learning, offering a principled, knowledge-driven alternative to data-driven tree construction.
\end{abstract}

\section{Introduction}
Tabular data is ubiquitous in various domains, such as healthcare, finance, and climate science. Traditional machine learning methods, such as logistic regression, gradient-boosted trees, and neural networks, have been widely used for tabular data classification. However, these methods often require a large amount of labeled data for training, which can be challenging to obtain in many real-world applications. This paper introduces a zero-shot approach using an LLM to build decision trees without actual data, relying solely on attribute information.

Decision trees are among the most widely used machine learning models due to their interpretability, simplicity, and effectiveness in classification and regression tasks~\cite{shwartz2022tabular,grinsztajn2022tree,borisov2022deep,mcelfresh2024neural}. Traditional methods for constructing decision trees rely heavily on access to labeled datasets, where attribute values and corresponding outcomes are used to recursively partition data based on criteria such as information gain or the Gini index~\cite{quinlan2014c4}. However, these approaches face significant limitations in scenarios where data is scarce, incomplete, or entirely unavailable, prompting the need for alternative methodologies.

Recent advances in large language models (LLMs) have demonstrated their remarkable ability to perform complex tasks in diverse domains by leveraging contextual understanding and knowledge encoded during pre-training~\cite{vaswani2017attention,brown2020language}. LLMs excel in zero-shot and few-shot learning paradigms, where minimal or no labeled examples are provided~\cite{raffel2020exploring,sanh2022multitask}, making them ideal candidates for addressing data-scarce challenges. This paper presents a novel zero-shot approach for constructing decision trees using an LLM, bypassing the need for traditional data-driven methods.

Our method leverages the contextual reasoning capabilities of LLMs to perform essential tasks needed for constructing decision trees. These tasks include discretizing continuous attributes and calculating conditional probabilities based on those attributes. This information allows us to recursively build the tree structure minimizing the Gini index. By framing the decision tree construction as a series of queries and responses to the LLM, we demonstrate that these models can effectively replicate the core steps of traditional algorithms without requiring explicit access to the underlying data.

Our zero-shot approach holds particular promise for applications where labeled datasets are difficult to obtain, such as in new or underrepresented domains, or for tasks involving sensitive data where privacy concerns limit access to detailed information. Furthermore, it paves the way for more interpretable uses of LLMs, as the resulting decision trees provide a transparent representation of the decision-making process~\cite{lipton2018mythos}.

The main contributions of this paper are as follows:
\begin{itemize}
    \item \textbf{Zero-Shot Learning for Decision Trees} We propose a decision tree algorithm that can be constructed without any labeled data, providing a foundation for knowledge-driven machine learning approaches.
    \item \textbf{Interpretable Models} We demonstrate how decision trees built using LLMs offer interpretable decision-making processes, addressing a crucial challenge in deploying AI systems for high-stakes applications.
    \item \textbf{Application to Low-Data Scenarios} We show how LLMs can effectively address the challenges of data-scarce environments by leveraging pre-existing knowledge, an approach especially valuable in domains with limited available data.
\end{itemize}

The remainder of this paper is organized as follows: Section~\ref{related_work} reviews related work on decision trees and LLM-based methodologies. Section~\ref{method} details the proposed zero-shot decision tree construction approach. Section~\ref{experiments} and ~\ref{results} presents experimental results, showcasing the performance and applicability of the method. Finally, Section~\ref{discussion} concludes with a discussion of the potential impact and future directions of this research.

\section{Related Work} \label{related_work}

\subsection{Classification with LLMs}
Transformer models~\cite{vaswani2017attention}, which form the backbone of most large language models, are pre-trained on vast amounts of text data and demonstrate remarkable versatility in generalizing to new tasks with minimal or no labeled examples. The architecture's ability to effectively leverage prior knowledge encoded within its parameters makes LLMs particularly attractive for few-shot learning scenarios~\cite{brown2020language,sanh2022multitask}.

\paragraph{Zero-Shot Classification}
Zero-shot learning has emerged as a promising approach for scenarios where labeled data is scarce or unavailable. This method leverages pre-trained models to predict new, unseen tasks without task-specific training data. 
Recent studies have shown the potential of zero-shot learning in various domains. Hegselmann et al.\cite{hegselmann2023tabllm} demonstrated that prompting LLMs with zero-shot or few-shot examples yields performance comparable to fine-tuned models in certain classification tasks. Similarly, pre-training on large tabular datasets further enhances an LLM's ability to generalize to unseen data, as shown by work in supervised tabular learning\cite{wen2024supervised}, unified table representation~\cite{yang2024unitabe}, and tabular Transformers~\cite{wang2022transtab}.

\paragraph{Supervised Learning} 
Applying deep learning to tabular data has received increasing attention in recent years, leading to several innovative transformer-based architectures. \cite{yin2020tabert} introduced \textit{TabERT}, a self-supervised model that improves learning from structured datasets through goals such as masked cell prediction and contrastive losses. Building on this, \cite{arik2021tabnet} proposed \textit{TabNet}, an attention-based framework specifically designed to capture sparse and meaningful feature interactions in tabular data. Similarly, \cite{somepalli2021saint,chen2023excelformer} developed \textit{SAINT} and \textit{ExcelFormer}, respectively, both of which propose novel self-attention mechanisms to model interactions in tabular datasets. \textit{MetaTree}~\cite{zhuang2024learning} uses transformers and meta-learning to recursively construct decision trees, mirroring classical greedy algorithms.
Other approaches, such as \cite{hollmann2022tabpfn}, introduced \textit{TabPFN}, a Bayesian neural network pre-trained on synthetic tabular datasets, demonstrating strong generalization capabilities across diverse tabular tasks.

\paragraph{Fine-Tuning} 
Another approach for classification is fine-tuning an LLM with a serialized representation of the tabular data and a description of the classification problem~\cite{dinh2022lift,hegselmann2023tabllm}. TapTap employs pre-training on a large corpus of real-world tabular data and can generate high-quality synthetic tables to improve the performance of prediction models~\cite{zhang2023generative}. 
To achieve this, TableGPT2~\cite{su2024tablegpt2} creates a dataset that includes various tabular datasets and tasks, such as classifications, for fine-tuning LLMs.
TableLlama~\cite{zhang2023tablellama} shows the potential of fine-tuning with multimodal models to address the challenges of table-based tasks.

\paragraph{Interpretable Classification in Machine Learning}
Despite advances in classification with machine learning methods, gradient-boosted tree ensembles remain the dominant choice for many practical applications due to their robustness, interpretability, and consistent performance across various tabular datasets. Classic methods such as XGBoost~\cite{chen2016xgboost} and LightGBM~\cite{ke2017lightgbm} continue to set the standard, often outperforming deep learning models in real-world scenarios~\cite{shwartz2022tabular,grinsztajn2022tree,borisov2022deep,mcelfresh2024neural}. 
Studies have highlighted the unique challenges deep learning models face with tabular data, including feature representation, small sample sizes, and overfitting, further reinforcing the advantages of traditional ensemble methods~\cite{borisov2022deep}. This persistent superiority emphasizes the need for hybrid approaches that combine the strengths of deep learning with the efficiency and interpretability of tree-based models~\cite{mcelfresh2024neural}.

\begin{figure}[t]
    \centering
    \includegraphics[width=0.8\linewidth]{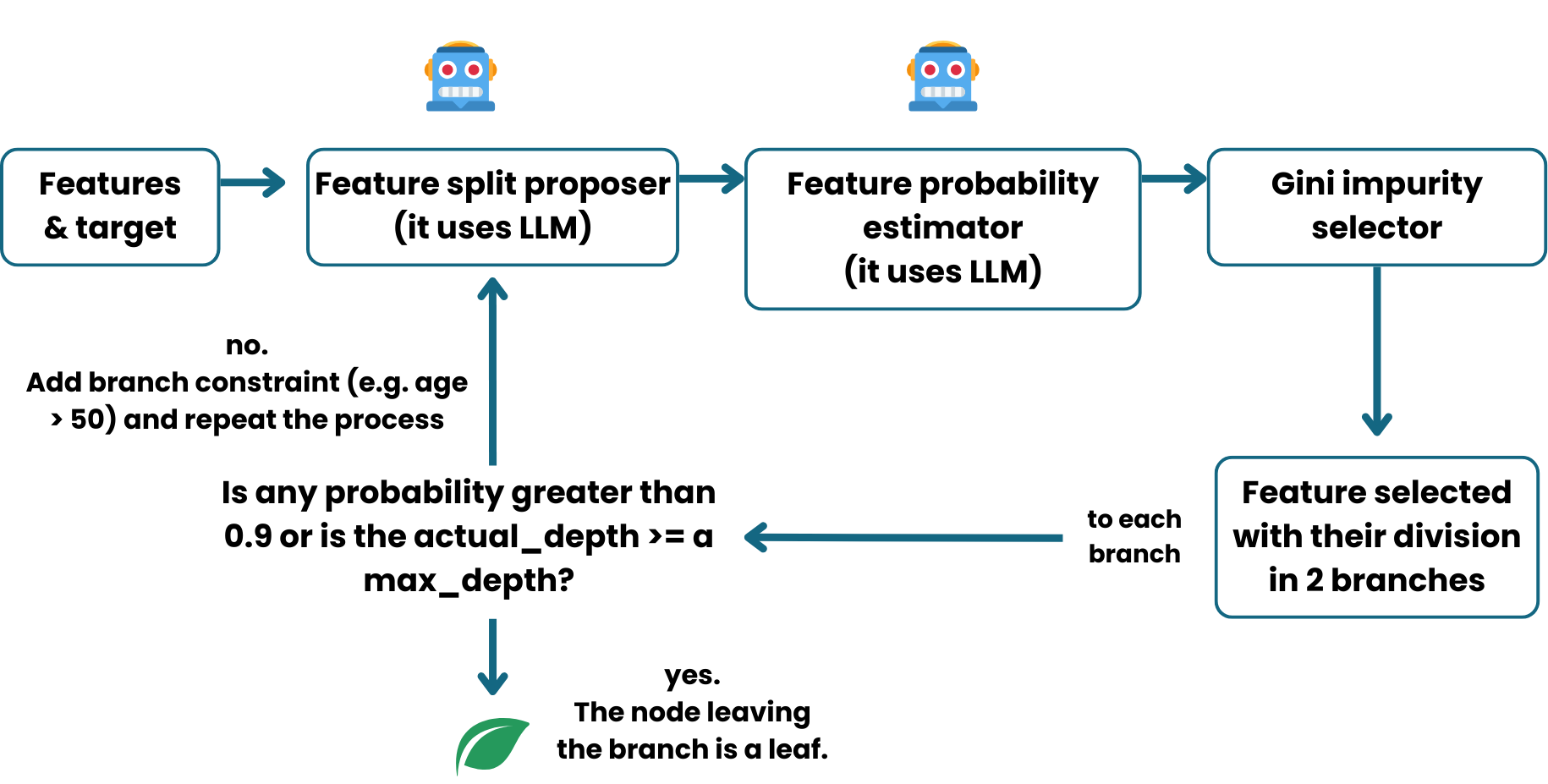}
    \caption{Overview of Zero-Shot DT Construction via LLM: The input is information from the features (it is not necessary to have data for this, only knowledge of the variables to be used).  The input is passed through the first 3 blocks that decide which feature will be used as a node to split the data into 2 branches. If the probabilities are greater than 0.9, or the depth is equal to or greater than the maximum depth, it is decided that the node is a leaf, and the label with the highest probability is given as a prediction; otherwise, the constraint imposed by the branch is added to a list of constraints, and iterates again, thus going down one level in the tree.}
    \label{fig:complete-diagram}
\end{figure}

\section{Methodology} \label{method}

This section introduces a novel methodology for constructing decision trees by utilizing the semantic understanding and contextual reasoning capabilities of LLMs. The process is designed for binary classification tasks and consists of three main steps: 1) attribute splitting, 2) probability estimation, and 3) Gini impurity minimization. The steps 1 and 2 utilize the reasoning abilities of the LLM to iteratively refine the structure of the tree. The following sections will describe the methodology in detail and explain the design choices made at each stage.

\subsection{Overview of the Process}
Figure~\ref{fig:complete-diagram} illustrates the methodology for constructing decision trees using LLMs designed for binary classification tasks. The inputs to this process are the names and descriptions of the attributes and the target variable, providing a clear understanding of their meaning. For example, consider a dataset where we aim to predict whether a patient has diabetes based on features like age, BMI, and glucose. levels.

The process begins with the Feature Split Proposer, where the LLM generates potential splits for the input features based on their descriptions. For instance, it might suggest splits such as "age > 50," "BMI > 30," or "glucose > 140," based on its understanding of how these features might relate to diabetes risk. Next, the Feature Probability Estimator uses the LLM to calculate the likelihood of the target outcome for each proposed split. For example, the model might estimate that patients with "glucose > 140" have an 85\% probability of having diabetes, while those with "BMI > 30" have a 70\% probability. These probabilities are derived using the LLM's ability to infer relationships between the features and the target variable from their descriptions.

The Gini Impurity Selector then evaluates the proposed splits and selects the one that minimizes the Gini impurity, ensuring that the resulting subsets are as homogeneous as possible. For instance, "glucose > 140" might be chosen as the first split because it creates subsets with a clearer distinction between diabetic and non-diabetic patients. 
This iterative process continues, adding constraints to further refine the tree structure. For example, after splitting on "glucose > 140," the next level might involve "age > 50" or "BMI > 30" to further segment the data. The tree grows until a leaf node is created, either when the probability of diabetes in a subset exceeds 0.9 or when the maximum tree depth is reached.

The prompts used for the LLM calls required significant refinement to achieve the desired results, particularly in the first two steps of the methodology: generating partitioning propositions for the attributes and estimating probabilities. During the development and experimentation process, several challenges arose. First, the proposed values or groupings for splitting a node into two groups often fail to satisfy the specified constraints. Second, the probability estimates tended to overemphasize certain classes, leading to imbalanced outcomes. Third, for categorical attributes, the suggested groups were not always mutually exclusive. To overcome these issues, the prompts were meticulously adjusted to provide explicit instructions, ensuring that the LLM generated valid splits, balanced probability estimates, and distinct groupings for categorical features. See Appendix Section~\ref{appendix_prompts} for more details on the prompts.




\subsection{Iterative Tree Construction}

The iterative tree-building process constructs a decision tree by systematically selecting and refining features based on their contribution to the problem context. This methodology attempts to make the tree partition the data optimally, respecting the defined constraints and taking advantage of the probabilistic reasoning of the LLM at each step.

To process the data through the tree, the names of the features and the categories of the categorical columns preferably have to be in lowercase, without spaces, special characters, etc. In particular, it is for some validations that are made in the algorithm.

The steps required to build the tree are described below.

\begin{enumerate}
    \item \textbf{Feature Splitting}: The process begins by identifying potential splits for each available feature. Candidate partitions are proposed for every feature based on the current problem context. These partitions aim to divide the feature space into subsets that align with the problem's decision boundaries. The proposals take into account the nature of the features (categorical or numerical) and ensure that the splits are both meaningful and feasible within the given constraints.

    \item \textbf{Probability Estimation}: Once the candidate partitions are generated, the next step is to estimate the probability of each label associated with each partition. These probabilities indicate the likelihood that an instance of the data will have the target attribute with a specific label for the subsets defined by the partitions. The estimates incorporate contextual information from the subtree being created, allowing the model to recognize which branches have been traversed. This step provides a quantitative basis for assessing the quality of each proposed split.

    \item \textbf{Optimal Split Selection}: To identify the best feature for splitting, the Gini impurity index is calculated for each candidate split. This metric measures the degree of impurity in the resulting subsets, with lower values indicating better separation of outcomes. The feature that minimizes Gini impurity is selected as the optimal split point for the current context. This ensures that the decision tree prioritizes features that maximize information gain at each step.
    \item \textbf{Branch Creation}: The selected feature and its corresponding split are applied to the current context, creating two branches. For each branch, the context is updated to reflect the new constraints introduced by the split. The process is then recursively repeated for each branch, treating the updated contexts as new problem instances.

    In cases where a branch reaches a predefined stopping criterion—such as a maximum depth, a high probability threshold for a single outcome, or the exhaustion of features—it is designated as a leaf node. The value of the leaf is determined based on the probabilities associated with the branch, selecting the most likely outcome.

    \item \textbf{Handling Feature Constraints}: As the tree grows, constraints on features are dynamically updated to ensure consistency and meaningful splits. The available categories are reduced for categorical features based on the chosen partition. If a feature's remaining categories become trivial (e.g., only one category remains), the feature is removed from consideration in subsequent splits. The upper or lower bounds are adjusted based on the split conditions for numerical features. This dynamic adjustment ensures that the tree construction respects feature-specific constraints and avoids redundant splits.

    \item \textbf{Termination and Final Tree Structure}: The recursive process continues until all branches reach a stopping criterion, in this case, reaching the maximum allowed depth or the node has a probability of 0.9 for one of the classes. The resulting tree is a hierarchical structure in which each node represents a decision based on a feature split, and each leaf represents a high confidence point. The tree is constructed to balance depth and precision, ensuring that it generalizes well to unseen instances while capturing the complexities of the problem space.
\end{enumerate}

This iterative framework produces a hierarchical structure with decision nodes representing feature splits and leaf nodes corresponding to high-confidence predictions.

Next, we detail the main steps involved in constructing a decision tree: 1) Attribute Splitting, 2) Probability Estimation, 3) Gini Impurity Minimization.

\subsection{Attribute Splitting}
The first step in constructing the decision tree is to generate meaningful splits based on the target variable being predicted for each attribute in the dataset. This process depends on the following inputs:

\begin{itemize}
    \item \textbf{Feature names}: Descriptive names that provide contextual information to the LLM. 
    \item \textbf{Feature types}: Indicating whether each feature is numerical or categorical. The possible categories are provided for categorical features.
    \item \textbf{Context and previous probabilities}: In cases where a subtree of a larger tree is being created, it is necessary to specify the probabilities of the root node of the subtree and the constraints met by the instances that reached that root.
\end{itemize}

All this information is passed to the language model via prompting, asking it to deliver a value, or groupings, that allow the data to be divided into 2 groups. The detailed prompts can be found in Appendix Section~\ref{appendix_numerical_split}.

\subsubsection{Numerical Attributes}

For numerical attributes, the LLM is tasked with proposing threshold values to partition the data. Each threshold splits the dataset into two groups: one with values less than or equal to the threshold and the other with values greater than the threshold. These thresholds are derived based on the LLM's contextual understanding and aim to maximize label homogeneity within each group.

For example, in a classification task with a numeric feature such as ``age,'' the LLM might propose a threshold of 35. This would result in two groups: ``age $\leq$ 35'' and ``age $>$ 35,'' if the split is expected to improve the separation of target labels. Additionally, the feature's data type (e.g., float or integer) is specified to ensure proper validation during the process.

A sample prompt for numerical attributes might look like this:

\begin{center}
\begin{minipage}{0.9\textwidth}
\textit{"Your role is to propose a single numeric value to divide the data into two groups based on the attribute \texttt{\{feature\}}..."}
\end{minipage}
\end{center}

\subsubsection{Categorical Attributes}

For categorical attributes, the LLM generates splits by dividing the available categories into two disjoint subsets. The objective is to maximize label separation while ensuring all categories are utilized in the split. The LLM leverages its semantic understanding to infer relationships between categories and generate contextually meaningful groupings.

For instance, given a categorical feature like ``color'' with categories $\{red, blue, green, yellow\}$, the LLM might propose splits such as $\{red, blue\}$ and $\{green, yellow\}$ based on perceived semantic or contextual similarities.

A typical prompt for categorical attributes is as follows:

\begin{center}
\begin{minipage}{0.9\textwidth}
\textit{"Your task is to use your knowledge to propose two disjoint groups of categories based on the categorical attribute \texttt{\{feature\}} with possible categories: \texttt{\{possible values\}}..."}
\end{minipage}
\end{center}

This approach ensures that both numerical and categorical splits are informed by the LLM's ability to reason contextually, resulting in splits that are both meaningful and effective for decision tree construction.

\subsection{Probability Estimation}

After proposing attribute splits, the next step involves estimating the probabilities of target labels for each resulting branch. The LLM performs this estimation through the following process:

\begin{enumerate}
    \item The two branches created by the split are presented to the LLM (e.g., ``values $\leq$ threshold'' and ``values $>$ threshold'' for numerical features).
    \item The LLM is tasked with estimating the likelihood of each target label for the instances in each branch.
    
    \begin{center}
    \begin{minipage}{0.9\textwidth}
    \textit{"Your task is to provide a rough estimate of the probabilities that a given instance belongs to the possible target classes based on: the characteristics, probabilities of the previous node, and the information from the new node split."}
    \end{minipage}
    \end{center}
\end{enumerate}

Using its knowledge base, the LLM infers probabilities by identifying general patterns, relationships, and contextual clues embedded in the instructions. This enables it to estimate probabilities even without access to explicit data distributions, leveraging its reasoning capabilities to make informed predictions~\cite{brown2020language}.

One critical refinement in this process involves addressing the tendency of the LLM to overestimate probabilities when limited information is available. During experimentation, this issue led to disproportionately high probabilities early in the tree construction, often resulting in trees with only one node. To mitigate this, the prompts explicitly instruct the LLM to avoid overconfidence and ensure more balanced probability estimates.

Estimating probabilities for each branch provides essential metrics to evaluate the quality of splits. These probabilities directly influence the calculation of metrics like Gini impurity, guiding the tree construction process toward optimal decisions. By incorporating contextual reasoning, the LLM ensures that the estimated probabilities are not only plausible but also aligned with the overarching classification objectives.

\subsection{Gini Impurity Minimization}

The Gini impurity index is used to evaluate the quality of proposed splits by quantifying the homogeneity of target labels in the resulting branches. The process follows these steps:

\begin{enumerate}
    \item \textbf{Compute the Gini impurity for each branch:}
    \[
    \text{Gini} = 1 - \sum_{i=1}^{C} p_i^2,
    \]
    where $p_i$ represents the probability of the $i$-th target label in the branch, and $C$ is the total number of target labels.
    
    \item \textbf{Aggregate the impurities of the two branches:}  
    The impurities are combined using the harmonic mean:
    \[
    \text{Harmonic Mean} = \frac{2 \cdot \text{Gini}_1 \cdot \text{Gini}_2}{\text{Gini}_1 + \text{Gini}_2},
    \]
    where $\text{Gini}_1$ and $\text{Gini}_2$ are the Gini impurities of the two branches.
\end{enumerate}

While the CART (Classification and Regression Tree)~\cite{breiman2017classification} method constructs decision trees by using a weighted average of the Gini impurities—where the weights are proportional to the number of instances in each branch—our approach operates without access to instance-level data. Instead, we employ the harmonic mean for aggregating impurity levels. 

The harmonic mean is deliberately chosen as it emphasizes a balance between the impurity levels of the two branches, penalizing splits that exhibit extreme imbalances. This ensures that the resulting splits are effective for both branches rather than disproportionately favoring one over the other. By guiding the decision tree construction process toward balanced and homogeneous partitions, this method enhances the overall quality of the tree, ensuring that target labels within each group are as uniform as possible.

\section{Experimental Design and Evaluation} \label{experiments}

To evaluate the proposed methodology for zero-shot decision tree construction using LLMS, we conducted a comparative analysis with two baseline approaches: (1) TabLLM zero-shot~\cite{hegselmann2023tabllm}, a state-of-the-art method for zero-shot tabular data classification using LLMs, and (2) traditional supervised Decision Trees trained using \texttt{Scikit-learn}~\cite{Pedregosa2012}. These methods were selected as benchmarks to assess the strengths and limitations of our approach, focusing on accuracy and Macro F1 scores across diverse classification tasks.

The experimental setup was designed to balance computational feasibility with model performance. For the traditional Decision Tree models, we varied the \texttt{max\_depth} parameter (set to 7, 5, and 3) to analyze its impact on predictive accuracy and interpretability. For the zero-shot methods, we leveraged the inherent capabilities of the LLM, GPT-4o Mini, to process tabular data without additional training, evaluating the computational trade-offs and advantages of leveraging LLMs for decision tree construction compared to traditional and state-of-the-art baselines.

To provide a comprehensive evaluation, we varied the amount of training data used for the supervised Decision Tree models, using subsets of 4, 8, 16, 32, and the full training dataset. This allowed us to assess the generalizability, efficiency, and applicability of the proposed method in scenarios with different levels of data availability.

Given that the proposed strategy is most effective in low-data environments, we carefully selected datasets from various domains, including health, finance, environment, and politics. This diversity allowed us to assess the adaptability of the proposed method in tackling a wide range of real-world problems. Additionally, we included the Presidential Approval dataset from 2024, which is beyond the training knowledge cutoff for GPT-4 models in October 2023~\cite{open_cutoff}. This provided a unique opportunity to test the generalizability and robustness of the zero-shot approach when dealing with entirely novel data.

By splitting each dataset into training and testing sets, we ensured meaningful comparisons with other strategies and evaluations against specific metrics. Training data was used to train models, while testing data was reserved to evaluate both data-intensive and zero-shot methods. This rigorous experimental design highlights the computational trade-offs, strengths, limitations, and practical applications of our proposed methodology.

\subsection{Baseline Methods}

\paragraph{TabLLM zero-shot} This baseline leverages the ability of LLMs to perform zero-shot classification directly on tabular data. TabLLM is designed to encode tabular data into text prompts, referred to as serialization, and combine it with a concise description of the classification problem. This enables LLMs to process tabular data as input text and leverage their inherent prior knowledge for classification without the need for training on the dataset.

\paragraph{Traditional Decision Trees} Decision trees were trained using \texttt{Scikit-learn} \cite{Pedregosa2012} to serve as a supervised learning baseline. These models rely on access to labeled training data and were evaluated using varying amounts of data to understand the impact of data availability on model performance. In our experiments, we used the \texttt{max\_depth} parameter of the decision tree model to control the maximum depth of the tree. 

\subsection{Dataset Selection}

To ensure a robust and diverse evaluation, we selected datasets from multiple domains, including health, finance, environment, and politics. This variety enabled us to assess the adaptability of our method to real-world problems. The datasets used in the study are as follows:

\paragraph{Diabetes\protect\footnote{\url{https://www.kaggle.com/datasets/uciml/pima-indians-diabetes-database}}} A health-related dataset focused on predicting the onset of diabetes based on individual measurements.

\paragraph{Credit\protect\footnote{\url{https://www.openml.org/search?type=data&status=active&id=31}}} A financial dataset containing information on past loans. The target variable predicts whether an individual is likely to repay the loan. 

\paragraph{Weather\protect\footnote{\url{https://www.kaggle.com/datasets/nikhil7280/weather-type-classification?select=weather_classification_data.csv}}} A dataset for binary weather prediction (sunny vs. rainy) based on environmental measurements such as temperature and humidity. The original dataset included multiple climates, but we transformed it into a binary classification problem by selecting two opposing climates. 
    
\paragraph{Presidential Approval\protect\footnote{\url{https://www.cepchile.cl/encuesta/encuesta-cep-n-92/}}} A dataset that predicts whether individuals approve or disapprove of Gabriel Boric's government based on demographic data (e.g., age, religion, education). This dataset is particularly significant because it contains recent data (2024) that the LLMs used in this study were not trained (The knowledge cutoff for GPT-4o models is October 2023~\cite{open_cutoff}), ensuring that the evaluation reflects the strategy's performance on unseen data. 

\paragraph{Hepatic damage\protect\footnote{\url{https://www.kaggle.com/datasets/fedesoriano/hepatitis-c-dataset}}} The dataset contains information on people who may have Hepatitis C, Fibrosis, Cirrhosis, be healthy patients, or be suspected. To transform the problem from a binary classification problem, the category column was transformed to the question:\textit{ Is there liver damage?} With this, we can transform the 3 disease categories into \textit{True} and the category of healthy people into \textit{False}. The rows of people who were in suspicion were eliminated.

\section{Results} \label{results}

Tables \ref{tab:resultados} and \ref{tab:resultados_2} summarize and compare the accuracy and macro F1 scores of the evaluated methods, including TabLLM zero-shot, Decision Trees (DTs) generated using LLMs in a zero-shot setting, and traditional decision trees\footnote{Here, we refer to a traditional decision tree (or traditional DT) as one obtained using the CART algorithm.}. These quantitative results, presented across a variety of datasets, provide a comprehensive overview of the methods' performance and are further complemented by a detailed discussion of the experimental results in subsequent sections.


\paragraph{Performance of TabLLM Zero Shot} TabLLM Zero Shot demonstrated moderate performance, excelling in simpler datasets like \emph{Weather}, where it achieved an accuracy and macro F1 score of 88\%. However, it struggled with datasets such as \emph{Diabetes} and \emph{Credit}, achieving lower accuracy and macro F1 scores of approximately 51\% and 45\%.

\paragraph{Performance of DT LLM Zero Shot} Our DT LLM Zero Shot generally outperformed TabLLM Zero Shot, particularly in datasets like \emph{Diabetes}, where a depth of 7 yielded an accuracy of 74\% and macro F1 of 70\%, significantly surpassing TabLLM. Nonetheless, its performance varied across datasets and tree depths, with notable declines in datasets like \emph{Weather} (accuracy: 50\%, macro F1: 35\%) and \emph{Credit} when tree depth was restricted to 3. Increased tree depth generally enhanced performance, as seen in the \emph{Diabetes} dataset, where accuracy rose from 49\% to 74\%. However, this trend was dataset-dependent; for instance, in \emph{Presidential Approval}, accuracy remained consistent ($\approx$ 40\%) across tree depths.

\paragraph{Traditional Decision Trees with Full Data} Traditional DTs trained on full datasets consistently outperformed zero-shot methods, achieving higher accuracy and macro F1 scores across most datasets. In the \emph{Hepatic Damage} dataset, traditional DTs achieved an accuracy of 93\% and macro F1 of 77\%, compared to 83\% and 63\% for DT LLM Zero Shot. Similarly, in the \emph{Credit} dataset, traditional DTs excelled with an accuracy of 76\% and macro F1 of 68\%.

\paragraph{Low-Data Scenarios.}  
In low-data scenarios (e.g., 4$-$32 shots), traditional DTs exhibited limitations, highlighting the competitive performance of LLM-based zero-shot approaches in resource-constrained environments. For example, with 4 shots,  traditional DTs performed comparably to DT LLM Zero Shot and were outperformed by our zero-shot method in datasets such as \emph{Diabetes}, \emph{Credit}, and \emph{Hepatic}, demonstrating the potential of LLM-based methods in handling limited data scenarios.

\begin{table}[t]
\centering
\begin{tabular}{l|cc|cc|cc}
\hline
\textbf{Methods} & \multicolumn{2}{c}{\textbf{Diabetes}} & \multicolumn{2}{c}{\textbf{Credit}} & \multicolumn{2}{c}{\textbf{Weather}} \\
 & Accuracy & Macro F1 & Accuracy & Macro F1 & Accuracy & Macro F1 \\
\hline
TabLLM zero-shot & .51 & .49 & .45 & .45 & .88 & .88\\
DT LLM zero-shot Depth=7 & .74 & .70 & .56 & .55 & .50 & .35  \\
DT LLM zero-shot Depth=5 & .75 & .71 & .50 & .50 & .50 & .35  \\
DT LLM zero-shot Depth=3 & .75 & .70 & .33 & .29 & .49 & .33  \\
\hline
Traditional DT Depth=7 (all data) & .73 & .71 & .71 & .67 & .94 & .94  \\
Traditional DT Depth=5 (all data) & .79 & .77 & .74 & .65 & .91 & .91  \\
Traditional DT Depth=3 (all data) & .76 & .73 & .76 & .68 & .86 & .86  \\
\hline
Traditional DT Depth=7 (4 shots) & .66 & .61 & .56 & .55 & .53 & .37  \\
Traditional DT Depth=7 (8 shots) & .59 & .58 & .59 & .58 & .69 & .69  \\
Traditional DT Depth=7 (16 shots) & .58 & .55 & .58 & .56 & .75 & .75  \\
Traditional DT Depth=7 (32 shots) & .62 & .58 & .60 & .55 & .84 & .84 \\
\hline
\end{tabular}
\caption{Comparison of accuracy and macro F1 scores for different methods applied to \emph{Diabetes}, \emph{Credit}, and \emph{Weather} datasets.}
\label{tab:resultados}
\end{table}
\begin{table}[ht]
\centering
\begin{tabular}{l|cc|cc}
\hline
\textbf{Methods} & \multicolumn{2}{c}{\textbf{Hepatic damage}} & \multicolumn{2}{c}{\textbf{Presidential Approval}} \\
 & Accuracy & Macro F1 & Accuracy & Macro F1 \\
\hline
TabLLM zero-shot & .88 & .47 & .66 & .52 \\
DT LLM zero-shot Depth=7 & .83 & .63 & .41 & .40 \\
DT LLM zero-shot Depth=5 & .83 & .63 & .42 & .41 \\
DT LLM zero-shot Depth=3 & .83 & .63 & .39 & .38 \\
\hline
Traditional DT Depth=7 (all data) & .92 & .75 & .71 & .62 \\
Traditional DT Depth=5 (all data) & .92 & .75 & .73 & .63 \\
Traditional DT Depth=3 (all data) & .93 & .77 & .70 & .46 \\
\hline
Traditional DT Depth=7 (4 shots) & .88 & .47 & .69 & .41 \\
Traditional DT Depth=7 (8 shots) & .88 & .47 & .61 & .44 \\
Traditional DT Depth=7 (16 shots) & .88 & .47 & .67 & .42\\
Traditional DT Depth=7 (32 shots) & .88 & .47  & .64 & .45 \\
\hline
\end{tabular}
\caption{Comparison of accuracy and macro F1 scores for different methods applied to \emph{Hepatic damage} and \emph{Presidential Approval} datasets.}
\label{tab:resultados_2}
\end{table}

\begin{figure}[t]
    \centering
    \includegraphics[width=1\linewidth]{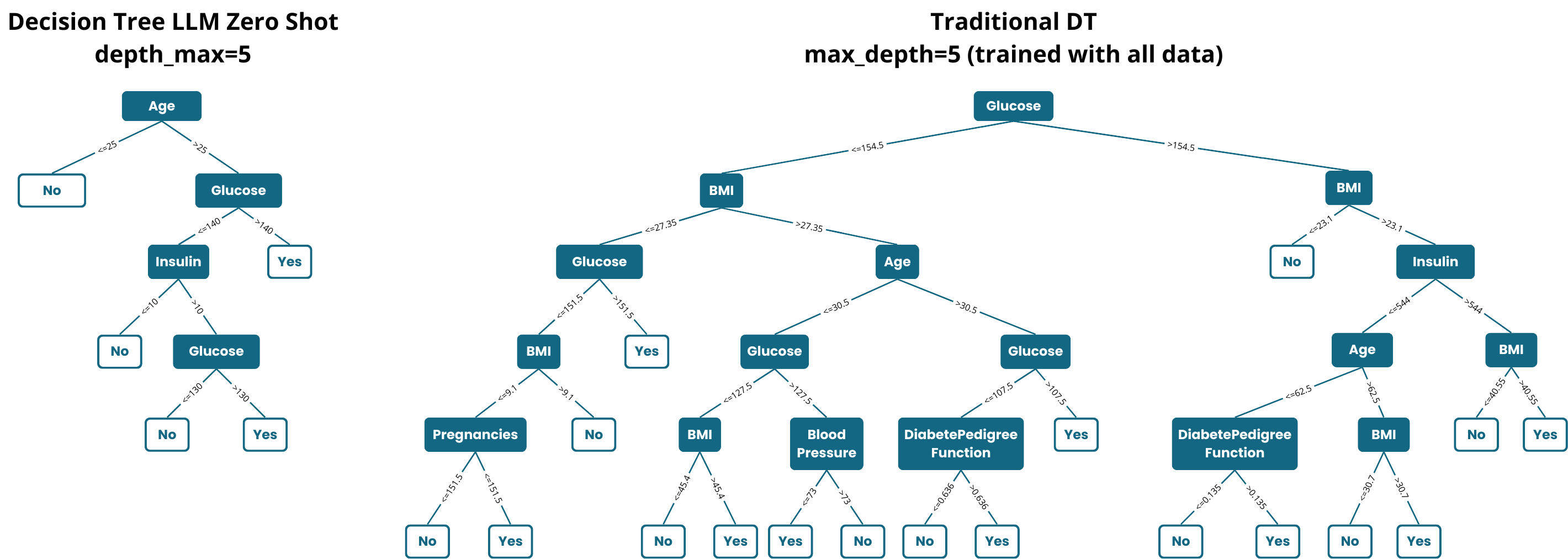}
        \caption{Comparison of decision trees: (Left) Zero-shot method with LLMs and (Right) CART algorithm for the \emph{Diabetes} dataset. These trees are simplified versions where branches leading to unanimous outcomes (\emph{Yes} or \emph{No}) are condensed to single nodes.}
    \label{fig:tree-fig}
\end{figure}

\subsection{Interpretability Analysis}
One of the main advantages of our proposed method its the interpretability of decision tress mixed with zero-shot training for low-resources scenarios. 
Figure \ref{fig:tree-fig} shows and example of a zero-shot tree constructed for the \emph{Diabetes} dataset, where their performances are similar, and compares it with a traditional tree trained on all the data. 


The zero-shot decision tree model, with a maximum depth of 5, offers a straightforward and interpretable structure, making it accessible to non-technical users. Its feature selection emphasizes domain-relevant variables like \texttt{Age}, \texttt{Glucose}, and \texttt{Insulin}, which aligns with user expectations and enhances trust in the model's decisions. The use of simple, intuitive decision thresholds, such as \(\texttt{Age} \leq 25\) or \(\texttt{Glucose} \leq 140\), further supports the interpretability of the model.

However, this approach sacrifices fine-grained distinctions in favor of simplicity, which may reduce accuracy and the model's ability to capture subtler patterns that traditional decision trees can identify. Additionally, the exclusion of features like \texttt{Pregnancies}, \texttt{Blood Pressure}, or \texttt{Diabetes Pedigree Function} suggests a potential trade-off between interpretability and comprehensiveness.

Overall, the zero-shot decision tree excels in interpretability but may lack robustness due to its simplified decision-making process, making it better suited for applications where ease of understanding takes precedence over detailed accuracy.


\section{Discussion} \label{discussion}

The proposed method of utilizing LLMs for zero-shot decision tree construction presents an innovative solution for low-data machine learning. This approach offers notable benefits in terms of efficiency, interpretability, and versatility across various scenarios. However, it also presents inherent challenges and trade-offs. This section delves into key aspects of the method, including computational efficiency, interpretability, and its applicability in low-data environments. We also examine its limitations, such as reliance on pre-trained knowledge, potential biases in LLM outputs, and the trade-off between interpretability and model complexity. Finally, we outline future directions aimed at improving and expanding this methodology.

\subsection{Advantages}

\paragraph{Computational Efficiency} While constructing decision trees using LLMs is computationally demanding, the resulting models are highly efficient during deployment. Unlike methods such as TabLLM, which incur significant computational costs at each inference, decision trees require only a one-time construction cost. This makes them particularly suitable for scenarios where models are deployed at scale or used for real-time decision-making. The reduced inference cost offsets the initial computational investment, offering long-term efficiency gains.

\paragraph{Explainability and Interpretability} Decision trees provide a clear and interpretable decision-making framework, which is a key advantage over black-box models. The ability to visualize and understand decision paths not only enhances user trust but also facilitates debugging and model validation. Furthermore, explainable models are often required in regulated industries, making decision trees an ideal choice. The structured nature of decision trees enables stakeholders to comprehend and justify decisions, fostering transparency and accountability.

\paragraph{Low-Data Scenarios} Leveraging LLMs for decision tree construction demonstrates significant potential in low-data environments. The pre-trained knowledge embedded in LLMs bridges the gap created by limited training data, enabling the creation of effective models even in challenging scenarios. This capability is particularly valuable in domains where data collection is expensive or impractical. By utilizing the vast knowledge encoded in LLMs, our approach provides a robust solution for addressing data scarcity while maintaining high model performance.

\subsection{Limitations}

\paragraph{LLM Bias} LLMs are trained on vast amounts of text data from diverse sources, inevitably including societal biases. These biases can be encoded into the knowledge base of the LLM and may inadvertently affect its outputs. When used for decision tree construction, these biases could manifest in selecting attributes, discretizing attribute values, or computation of probabilities, potentially leading to models that reflect and perpetuate these biases. For example, the LLM might prioritize certain features based on its training data rather than their relevance to the problem domain, inadvertently embedding stereotypical or unfair assumptions into the model.
Careful monitoring of the LLM's outputs and evaluation of resulting models are needed to detect and correct this bias. Techniques such as auditing the generated trees for fairness, analyzing the distributions of decisions across sensitive attributes, and leveraging fairness-aware metrics can help identify and address bias. Additionally, incorporating domain expertise to cross-validate the model's decisions can provide a safeguard against unintentional perpetuation of harmful biases \cite{mehrabi2021survey}.

\paragraph{Dependency on Pre-trained Knowledge} The proposed method relies heavily on the pre-trained knowledge embedded in the LLM. While this allows for zero-shot decision tree construction, it also means that the quality and accuracy of the trees are inherently limited by the breadth and depth of the LLM's training data~\cite{bommasani2021opportunities}. If the LLM lacks sufficient knowledge about a particular domain or exhibits inaccuracies in its responses, this could compromise the performance of the resulting decision tree. One potential mitigation strategy is fine-tuning the LLM on domain-specific data by tailoring the model's knowledge to the specific problem context.

\paragraph{Interpretability vs. LLM Complexity} One of the key advantages of decision trees is their interpretability. However, the involvement of an LLM introduces a layer of complexity that may not be fully transparent. For instance, the rationale behind the LLM's decisions during attribute selection or probability assignment might not always be explainable, potentially reducing the model's overall transparency. Achieving a balance between the clarity of decision trees and the inherent complexity of LLM operations is essential and raises important research questions.

\subsection{Future Work}

\paragraph{Interactive Tree Refinement} A promising extension to the proposed zero-shot decision tree construction approach involves introducing an interactive refinement process. This process allows human experts to refine or validate the tree structure iteratively during its construction. The LLM would propose initial splits and decisions, while the user could provide feedback on splits, override decisions based on domain knowledge or context, and suggest additional attributes or considerations the LLM might have overlooked.

\paragraph{Exploration of Hybrid Methods} While the DT LLM zero-shot method shows significant potential for low-data environments, traditional decision trees remain the more reliable choice in scenarios where large amounts of data are available. Future work should focus on optimizing the zero-shot approach for better generalization across diverse datasets and exploring hybrid methods that combine the strengths of both traditional and zero-shot techniques.

\paragraph{Domain-Specific Fine-Tuning} While the LLM’s general knowledge proved useful, fine-tuning the model on domain-specific data could significantly enhance the accuracy and relevance of the constructed trees. Future work could explore techniques for targeted adaptation of LLMs to specific fields, ensuring alignment with the nuances of specialized datasets.

\paragraph{Fairness and Bias Mitigation} Biases embedded in the LLM can influence the construction of decision trees. Future research could integrate fairness-aware algorithms into the tree-building process, ensuring that attribute selection, discretization, and splitting decisions account for fairness considerations and regulatory requirements.

\section{Conclusion}

This paper introduces a zero-shot method for constructing decision trees using Large Language Models, leveraging their pre-trained knowledge without the need for labeled data. This approach combines the strengths of LLMs with decision tree algorithms, offering a balance between efficiency, interpretability, and transparency. Zero-shot methods show promise for rapid prototyping and decision-making in resource-constrained environments, particularly in scenarios where data scarcity limits the use of traditional methods.

While the DT LLM zero-shot method demonstrates potential, traditional decision trees trained on full datasets consistently offer more robust and reliable performance. The results indicate that the DT LLM zero-shot method performs competitively in low-data environments where traditional models may struggle or require substantial data preprocessing. 

The performance of the zero-shot method is sensitive to the depth of the tree. Generally, deeper trees enhance accuracy, demonstrating the method's ability to scale with increased complexity. However, zero-shot approaches may still need further optimization to effectively manage the intricacies of diverse datasets. This indicates that while zero-shot techniques can be valuable tools, they might not be universally applicable or as reliable as traditional methods, especially when dealing with datasets that have significant complexity.

\section{Adverse Impact Statement}
The use of LLMs in the zero-shot model introduces inherent risks related to bias and discrimination, as these models are trained on vast amounts of data that may contain historical and systemic biases.  This may result in models that not only reflect but also reinforce these biases.

\bibliographystyle{ACM-Reference-Format}
\bibliography{Bib}


\begin{thebibliography}{31}


\ifx \showCODEN    \undefined \def \showCODEN     #1{\unskip}     \fi
\ifx \showDOI      \undefined \def \showDOI       #1{#1}\fi
\ifx \showISBNx    \undefined \def \showISBNx     #1{\unskip}     \fi
\ifx \showISBNxiii \undefined \def \showISBNxiii  #1{\unskip}     \fi
\ifx \showISSN     \undefined \def \showISSN      #1{\unskip}     \fi
\ifx \showLCCN     \undefined \def \showLCCN      #1{\unskip}     \fi
\ifx \shownote     \undefined \def \shownote      #1{#1}          \fi
\ifx \showarticletitle \undefined \def \showarticletitle #1{#1}   \fi
\ifx \showURL      \undefined \def \showURL       {\relax}        \fi
\providecommand\bibfield[2]{#2}
\providecommand\bibinfo[2]{#2}
\providecommand\natexlab[1]{#1}
\providecommand\showeprint[2][]{arXiv:#2}

\bibitem[Arik and Pfister(2021)]%
        {arik2021tabnet}
\bibfield{author}{\bibinfo{person}{Sercan~{\"O} Arik} {and}
  \bibinfo{person}{Tomas Pfister}.} \bibinfo{year}{2021}\natexlab{}.
\newblock \showarticletitle{Tabnet: Attentive interpretable tabular learning}.
  In \bibinfo{booktitle}{\emph{Proceedings of the AAAI conference on artificial
  intelligence}}, Vol.~\bibinfo{volume}{35}. \bibinfo{pages}{6679--6687}.
\newblock


\bibitem[Bommasani et~al\mbox{.}(2021)]%
        {bommasani2021opportunities}
\bibfield{author}{\bibinfo{person}{Rishi Bommasani}, \bibinfo{person}{Drew~A
  Hudson}, \bibinfo{person}{Ehsan Adeli}, \bibinfo{person}{Russ Altman},
  \bibinfo{person}{Simran Arora}, \bibinfo{person}{Sydney von Arx},
  \bibinfo{person}{Michael~S Bernstein}, \bibinfo{person}{Jeannette Bohg},
  \bibinfo{person}{Antoine Bosselut}, \bibinfo{person}{Emma Brunskill},
  {et~al\mbox{.}}} \bibinfo{year}{2021}\natexlab{}.
\newblock \showarticletitle{On the opportunities and risks of foundation
  models}.
\newblock \bibinfo{journal}{\emph{arXiv preprint arXiv:2108.07258}}
  (\bibinfo{year}{2021}).
\newblock


\bibitem[Borisov et~al\mbox{.}(2022)]%
        {borisov2022deep}
\bibfield{author}{\bibinfo{person}{Vadim Borisov}, \bibinfo{person}{Tobias
  Leemann}, \bibinfo{person}{Kathrin Se{\ss}ler}, \bibinfo{person}{Johannes
  Haug}, \bibinfo{person}{Martin Pawelczyk}, {and} \bibinfo{person}{Gjergji
  Kasneci}.} \bibinfo{year}{2022}\natexlab{}.
\newblock \showarticletitle{Deep neural networks and tabular data: A survey}.
\newblock \bibinfo{journal}{\emph{IEEE transactions on neural networks and
  learning systems}} (\bibinfo{year}{2022}).
\newblock


\bibitem[Breiman(2017)]%
        {breiman2017classification}
\bibfield{author}{\bibinfo{person}{Leo Breiman}.}
  \bibinfo{year}{2017}\natexlab{}.
\newblock \bibinfo{booktitle}{\emph{Classification and regression trees}}.
\newblock \bibinfo{publisher}{Routledge}.
\newblock


\bibitem[Brown et~al\mbox{.}(2020)]%
        {brown2020language}
\bibfield{author}{\bibinfo{person}{Tom Brown}, \bibinfo{person}{Benjamin Mann},
  \bibinfo{person}{Nick Ryder}, \bibinfo{person}{Melanie Subbiah},
  \bibinfo{person}{Jared~D Kaplan}, \bibinfo{person}{Prafulla Dhariwal},
  \bibinfo{person}{Arvind Neelakantan}, \bibinfo{person}{Pranav Shyam},
  \bibinfo{person}{Girish Sastry}, \bibinfo{person}{Amanda Askell},
  {et~al\mbox{.}}} \bibinfo{year}{2020}\natexlab{}.
\newblock \showarticletitle{Language models are few-shot learners}.
\newblock \bibinfo{journal}{\emph{Advances in neural information processing
  systems}}  \bibinfo{volume}{33} (\bibinfo{year}{2020}),
  \bibinfo{pages}{1877--1901}.
\newblock


\bibitem[Chen et~al\mbox{.}(2023)]%
        {chen2023excelformer}
\bibfield{author}{\bibinfo{person}{Jintai Chen}, \bibinfo{person}{Jiahuan Yan},
  \bibinfo{person}{Qiyuan Chen}, \bibinfo{person}{Danny~Ziyi Chen},
  \bibinfo{person}{Jian Wu}, {and} \bibinfo{person}{Jimeng Sun}.}
  \bibinfo{year}{2023}\natexlab{}.
\newblock \showarticletitle{Excelformer: A neural network surpassing gbdts on
  tabular data}.
\newblock \bibinfo{journal}{\emph{arXiv preprint arXiv:2301.02819}}
  (\bibinfo{year}{2023}).
\newblock


\bibitem[Chen and Guestrin(2016)]%
        {chen2016xgboost}
\bibfield{author}{\bibinfo{person}{Tianqi Chen} {and} \bibinfo{person}{Carlos
  Guestrin}.} \bibinfo{year}{2016}\natexlab{}.
\newblock \showarticletitle{Xgboost: A scalable tree boosting system}. In
  \bibinfo{booktitle}{\emph{Proceedings of the 22nd acm sigkdd international
  conference on knowledge discovery and data mining}}.
  \bibinfo{pages}{785--794}.
\newblock


\bibitem[Dinh et~al\mbox{.}(2022)]%
        {dinh2022lift}
\bibfield{author}{\bibinfo{person}{Tuan Dinh}, \bibinfo{person}{Yuchen Zeng},
  \bibinfo{person}{Ruisu Zhang}, \bibinfo{person}{Ziqian Lin},
  \bibinfo{person}{Michael Gira}, \bibinfo{person}{Shashank Rajput},
  \bibinfo{person}{Jy-yong Sohn}, \bibinfo{person}{Dimitris Papailiopoulos},
  {and} \bibinfo{person}{Kangwook Lee}.} \bibinfo{year}{2022}\natexlab{}.
\newblock \showarticletitle{Lift: Language-interfaced fine-tuning for
  non-language machine learning tasks}.
\newblock \bibinfo{journal}{\emph{Advances in Neural Information Processing
  Systems}}  \bibinfo{volume}{35} (\bibinfo{year}{2022}),
  \bibinfo{pages}{11763--11784}.
\newblock


\bibitem[Grinsztajn et~al\mbox{.}(2022)]%
        {grinsztajn2022tree}
\bibfield{author}{\bibinfo{person}{L{\'e}o Grinsztajn},
  \bibinfo{person}{Edouard Oyallon}, {and} \bibinfo{person}{Ga{\"e}l
  Varoquaux}.} \bibinfo{year}{2022}\natexlab{}.
\newblock \showarticletitle{Why do tree-based models still outperform deep
  learning on typical tabular data?}
\newblock \bibinfo{journal}{\emph{Advances in neural information processing
  systems}}  \bibinfo{volume}{35} (\bibinfo{year}{2022}),
  \bibinfo{pages}{507--520}.
\newblock


\bibitem[Hegselmann et~al\mbox{.}(2023)]%
        {hegselmann2023tabllm}
\bibfield{author}{\bibinfo{person}{Stefan Hegselmann},
  \bibinfo{person}{Alejandro Buendia}, \bibinfo{person}{Hunter Lang},
  \bibinfo{person}{Monica Agrawal}, \bibinfo{person}{Xiaoyi Jiang}, {and}
  \bibinfo{person}{David Sontag}.} \bibinfo{year}{2023}\natexlab{}.
\newblock \showarticletitle{Tabllm: Few-shot classification of tabular data
  with large language models}. In \bibinfo{booktitle}{\emph{International
  Conference on Artificial Intelligence and Statistics}}. PMLR,
  \bibinfo{pages}{5549--5581}.
\newblock


\bibitem[Hollmann et~al\mbox{.}(2022)]%
        {hollmann2022tabpfn}
\bibfield{author}{\bibinfo{person}{Noah Hollmann}, \bibinfo{person}{Samuel
  M{\"u}ller}, \bibinfo{person}{Katharina Eggensperger}, {and}
  \bibinfo{person}{Frank Hutter}.} \bibinfo{year}{2022}\natexlab{}.
\newblock \showarticletitle{Tabpfn: A transformer that solves small tabular
  classification problems in a second}.
\newblock \bibinfo{journal}{\emph{arXiv preprint arXiv:2207.01848}}
  (\bibinfo{year}{2022}).
\newblock


\bibitem[Ke et~al\mbox{.}(2017)]%
        {ke2017lightgbm}
\bibfield{author}{\bibinfo{person}{Guolin Ke}, \bibinfo{person}{Qi Meng},
  \bibinfo{person}{Thomas Finley}, \bibinfo{person}{Taifeng Wang},
  \bibinfo{person}{Wei Chen}, \bibinfo{person}{Weidong Ma},
  \bibinfo{person}{Qiwei Ye}, {and} \bibinfo{person}{Tie-Yan Liu}.}
  \bibinfo{year}{2017}\natexlab{}.
\newblock \showarticletitle{Lightgbm: A highly efficient gradient boosting
  decision tree}.
\newblock \bibinfo{journal}{\emph{Advances in neural information processing
  systems}}  \bibinfo{volume}{30} (\bibinfo{year}{2017}).
\newblock


\bibitem[Lipton(2018)]%
        {lipton2018mythos}
\bibfield{author}{\bibinfo{person}{Zachary~C Lipton}.}
  \bibinfo{year}{2018}\natexlab{}.
\newblock \showarticletitle{The mythos of model interpretability: In machine
  learning, the concept of interpretability is both important and slippery.}
\newblock \bibinfo{journal}{\emph{Queue}} \bibinfo{volume}{16},
  \bibinfo{number}{3} (\bibinfo{year}{2018}), \bibinfo{pages}{31--57}.
\newblock


\bibitem[McElfresh et~al\mbox{.}(2024)]%
        {mcelfresh2024neural}
\bibfield{author}{\bibinfo{person}{Duncan McElfresh}, \bibinfo{person}{Sujay
  Khandagale}, \bibinfo{person}{Jonathan Valverde}, \bibinfo{person}{Vishak
  Prasad~C}, \bibinfo{person}{Ganesh Ramakrishnan}, \bibinfo{person}{Micah
  Goldblum}, {and} \bibinfo{person}{Colin White}.}
  \bibinfo{year}{2024}\natexlab{}.
\newblock \showarticletitle{When do neural nets outperform boosted trees on
  tabular data?}
\newblock \bibinfo{journal}{\emph{Advances in Neural Information Processing
  Systems}}  \bibinfo{volume}{36} (\bibinfo{year}{2024}).
\newblock


\bibitem[Mehrabi et~al\mbox{.}(2021)]%
        {mehrabi2021survey}
\bibfield{author}{\bibinfo{person}{Ninareh Mehrabi}, \bibinfo{person}{Fred
  Morstatter}, \bibinfo{person}{Nripsuta Saxena}, \bibinfo{person}{Kristina
  Lerman}, {and} \bibinfo{person}{Aram Galstyan}.}
  \bibinfo{year}{2021}\natexlab{}.
\newblock \showarticletitle{A survey on bias and fairness in machine learning}.
\newblock \bibinfo{journal}{\emph{ACM computing surveys (CSUR)}}
  \bibinfo{volume}{54}, \bibinfo{number}{6} (\bibinfo{year}{2021}),
  \bibinfo{pages}{1--35}.
\newblock


\bibitem[OpenAI(2023)]%
        {open_cutoff}
\bibfield{author}{\bibinfo{person}{OpenAI}.} \bibinfo{year}{2023}\natexlab{}.
\newblock \bibinfo{title}{Models}.
\newblock
  \bibinfo{howpublished}{\url{https://platform.openai.com/docs/models\#gpt-4o}}.
\newblock
\newblock
\shownote{Accessed: January 20, 2024}.


\bibitem[Pedregosa et~al\mbox{.}(2012)]%
        {Pedregosa2012}
\bibfield{author}{\bibinfo{person}{Fabian Pedregosa}, \bibinfo{person}{Gaël
  Varoquaux}, \bibinfo{person}{Alexandre Gramfort}, \bibinfo{person}{Vincent
  Michel}, \bibinfo{person}{Bertrand Thirion}, \bibinfo{person}{Olivier
  Grisel}, \bibinfo{person}{Mathieu Blondel}, \bibinfo{person}{Peter
  Prettenhofer}, \bibinfo{person}{Ron Weiss}, \bibinfo{person}{Vincent
  Dubourg}, \bibinfo{person}{Jake Vanderplas}, \bibinfo{person}{Alexandre
  Passos}, \bibinfo{person}{David Cournapeau}, \bibinfo{person}{Matthieu
  Brucher}, \bibinfo{person}{Matthieu Perrot}, {and} \bibinfo{person}{Édouard
  Duchesnay}.} \bibinfo{year}{2012}\natexlab{}.
\newblock \showarticletitle{{Scikit-learn: Machine Learning in Python}}.
\newblock \bibinfo{journal}{\emph{Journal of Machine Learning Research}}
  \bibinfo{volume}{12} (\bibinfo{year}{2012}), \bibinfo{pages}{2825--2830}.
\newblock
\showISBNx{1532-4435}
\showISSN{15324435}
\urldef\tempurl%
\url{https://doi.org/10.1007/s13398-014-0173-7.2}
\showDOI{\tempurl}


\bibitem[Quinlan(2014)]%
        {quinlan2014c4}
\bibfield{author}{\bibinfo{person}{J~Ross Quinlan}.}
  \bibinfo{year}{2014}\natexlab{}.
\newblock \bibinfo{booktitle}{\emph{C4. 5: programs for machine learning}}.
\newblock \bibinfo{publisher}{Elsevier}.
\newblock


\bibitem[Raffel et~al\mbox{.}(2020)]%
        {raffel2020exploring}
\bibfield{author}{\bibinfo{person}{Colin Raffel}, \bibinfo{person}{Noam
  Shazeer}, \bibinfo{person}{Adam Roberts}, \bibinfo{person}{Katherine Lee},
  \bibinfo{person}{Sharan Narang}, \bibinfo{person}{Michael Matena},
  \bibinfo{person}{Yanqi Zhou}, \bibinfo{person}{Wei Li}, {and}
  \bibinfo{person}{Peter~J Liu}.} \bibinfo{year}{2020}\natexlab{}.
\newblock \showarticletitle{Exploring the limits of transfer learning with a
  unified text-to-text transformer}.
\newblock \bibinfo{journal}{\emph{Journal of machine learning research}}
  \bibinfo{volume}{21}, \bibinfo{number}{140} (\bibinfo{year}{2020}),
  \bibinfo{pages}{1--67}.
\newblock


\bibitem[Sanh et~al\mbox{.}(2022)]%
        {sanh2022multitask}
\bibfield{author}{\bibinfo{person}{Victor Sanh}, \bibinfo{person}{Albert
  Webson}, \bibinfo{person}{Colin Raffel}, \bibinfo{person}{Stephen Bach},
  \bibinfo{person}{Lintang Sutawika}, \bibinfo{person}{Zaid Alyafeai},
  \bibinfo{person}{Antoine Chaffin}, \bibinfo{person}{Arnaud Stiegler},
  \bibinfo{person}{Arun Raja}, \bibinfo{person}{Manan Dey}, {et~al\mbox{.}}}
  \bibinfo{year}{2022}\natexlab{}.
\newblock \showarticletitle{Multitask prompted training enables zero-shot task
  generalization}. In \bibinfo{booktitle}{\emph{International Conference on
  Learning Representations}}.
\newblock


\bibitem[Shwartz-Ziv and Armon(2022)]%
        {shwartz2022tabular}
\bibfield{author}{\bibinfo{person}{Ravid Shwartz-Ziv} {and}
  \bibinfo{person}{Amitai Armon}.} \bibinfo{year}{2022}\natexlab{}.
\newblock \showarticletitle{Tabular data: Deep learning is not all you need}.
\newblock \bibinfo{journal}{\emph{Information Fusion}}  \bibinfo{volume}{81}
  (\bibinfo{year}{2022}), \bibinfo{pages}{84--90}.
\newblock


\bibitem[Somepalli et~al\mbox{.}(2021)]%
        {somepalli2021saint}
\bibfield{author}{\bibinfo{person}{Gowthami Somepalli}, \bibinfo{person}{Micah
  Goldblum}, \bibinfo{person}{Avi Schwarzschild}, \bibinfo{person}{C~Bayan
  Bruss}, {and} \bibinfo{person}{Tom Goldstein}.}
  \bibinfo{year}{2021}\natexlab{}.
\newblock \showarticletitle{Saint: Improved neural networks for tabular data
  via row attention and contrastive pre-training}.
\newblock \bibinfo{journal}{\emph{arXiv preprint arXiv:2106.01342}}
  (\bibinfo{year}{2021}).
\newblock


\bibitem[Su et~al\mbox{.}(2024)]%
        {su2024tablegpt2}
\bibfield{author}{\bibinfo{person}{Aofeng Su}, \bibinfo{person}{Aowen Wang},
  \bibinfo{person}{Chao Ye}, \bibinfo{person}{Chen Zhou}, \bibinfo{person}{Ga
  Zhang}, \bibinfo{person}{Guangcheng Zhu}, \bibinfo{person}{Haobo Wang},
  \bibinfo{person}{Haokai Xu}, \bibinfo{person}{Hao Chen},
  \bibinfo{person}{Haoze Li}, {et~al\mbox{.}}} \bibinfo{year}{2024}\natexlab{}.
\newblock \showarticletitle{Tablegpt2: A large multimodal model with tabular
  data integration}.
\newblock \bibinfo{journal}{\emph{arXiv preprint arXiv:2411.02059}}
  (\bibinfo{year}{2024}).
\newblock


\bibitem[Vaswani(2017)]%
        {vaswani2017attention}
\bibfield{author}{\bibinfo{person}{A Vaswani}.}
  \bibinfo{year}{2017}\natexlab{}.
\newblock \showarticletitle{Attention is all you need}.
\newblock \bibinfo{journal}{\emph{Advances in Neural Information Processing
  Systems}} (\bibinfo{year}{2017}).
\newblock


\bibitem[Wang and Sun(2022)]%
        {wang2022transtab}
\bibfield{author}{\bibinfo{person}{Zifeng Wang} {and} \bibinfo{person}{Jimeng
  Sun}.} \bibinfo{year}{2022}\natexlab{}.
\newblock \showarticletitle{Transtab: Learning transferable tabular
  transformers across tables}.
\newblock \bibinfo{journal}{\emph{Advances in Neural Information Processing
  Systems}}  \bibinfo{volume}{35} (\bibinfo{year}{2022}),
  \bibinfo{pages}{2902--2915}.
\newblock


\bibitem[Wen et~al\mbox{.}(2024)]%
        {wen2024supervised}
\bibfield{author}{\bibinfo{person}{Xumeng Wen}, \bibinfo{person}{Han Zhang},
  \bibinfo{person}{Shun Zheng}, \bibinfo{person}{Wei Xu}, {and}
  \bibinfo{person}{Jiang Bian}.} \bibinfo{year}{2024}\natexlab{}.
\newblock \showarticletitle{From supervised to generative: A novel paradigm for
  tabular deep learning with large language models}. In
  \bibinfo{booktitle}{\emph{Proceedings of the 30th ACM SIGKDD Conference on
  Knowledge Discovery and Data Mining}}. \bibinfo{pages}{3323--3333}.
\newblock


\bibitem[Yang et~al\mbox{.}(2024)]%
        {yang2024unitabe}
\bibfield{author}{\bibinfo{person}{Yazheng Yang}, \bibinfo{person}{Yuqi Wang},
  \bibinfo{person}{Guang Liu}, \bibinfo{person}{Ledell Wu}, {and}
  \bibinfo{person}{Qi Liu}.} \bibinfo{year}{2024}\natexlab{}.
\newblock \showarticletitle{Unitabe: A universal pretraining protocol for
  tabular foundation model in data science}. In \bibinfo{booktitle}{\emph{The
  Twelfth International Conference on Learning Representations}}.
\newblock


\bibitem[Yin et~al\mbox{.}(2020)]%
        {yin2020tabert}
\bibfield{author}{\bibinfo{person}{Pengcheng Yin}, \bibinfo{person}{Graham
  Neubig}, \bibinfo{person}{Wen-tau Yih}, {and} \bibinfo{person}{Sebastian
  Riedel}.} \bibinfo{year}{2020}\natexlab{}.
\newblock \showarticletitle{TaBERT: Pretraining for Joint Understanding of
  Textual and Tabular Data}. In \bibinfo{booktitle}{\emph{Proceedings of the
  58th Annual Meeting of the Association for Computational Linguistics}}.
  \bibinfo{pages}{8413--8426}.
\newblock


\bibitem[Zhang et~al\mbox{.}(2023a)]%
        {zhang2023generative}
\bibfield{author}{\bibinfo{person}{Tianping Zhang}, \bibinfo{person}{Shaowen
  Wang}, \bibinfo{person}{Shuicheng Yan}, \bibinfo{person}{Jian Li}, {and}
  \bibinfo{person}{Qian Liu}.} \bibinfo{year}{2023}\natexlab{a}.
\newblock \showarticletitle{Generative table pre-training empowers models for
  tabular prediction}.
\newblock \bibinfo{journal}{\emph{arXiv preprint arXiv:2305.09696}}
  (\bibinfo{year}{2023}).
\newblock


\bibitem[Zhang et~al\mbox{.}(2023b)]%
        {zhang2023tablellama}
\bibfield{author}{\bibinfo{person}{Tianshu Zhang}, \bibinfo{person}{Xiang Yue},
  \bibinfo{person}{Yifei Li}, {and} \bibinfo{person}{Huan Sun}.}
  \bibinfo{year}{2023}\natexlab{b}.
\newblock \showarticletitle{Tablellama: Towards open large generalist models
  for tables}.
\newblock \bibinfo{journal}{\emph{arXiv preprint arXiv:2311.09206}}
  (\bibinfo{year}{2023}).
\newblock


\bibitem[Zhuang et~al\mbox{.}(2024)]%
        {zhuang2024learning}
\bibfield{author}{\bibinfo{person}{Yufan Zhuang}, \bibinfo{person}{Liyuan Liu},
  \bibinfo{person}{Chandan Singh}, \bibinfo{person}{Jingbo Shang}, {and}
  \bibinfo{person}{Jianfeng Gao}.} \bibinfo{year}{2024}\natexlab{}.
\newblock \showarticletitle{Learning a Decision Tree Algorithm with
  Transformers}.
\newblock \bibinfo{journal}{\emph{arXiv preprint arXiv:2402.03774}}
  (\bibinfo{year}{2024}).
\newblock


\end{thebibliography}

\newpage
\section*{Appendix}

\section{Prompts}
\label{appendix_prompts}

This section presents the prompts in each step requiring an LLM call. Notably, two prompts are used for the three steps involving LLMs: one instructs the LLM to perform a task, and the other extracts its response, the parser. 
The approach was refined to request direct, concise answers instead of detailed analysis to improve efficiency and reduce computational costs.

Each prompt consists of three parts: the system message, the assistant's response, and the user's request.

\subsection{Attribute Splitting Proposals}

\subsubsection{Get Numerical Features Splits}
\label{appendix_numerical_split}

System message:

\begin{lstlisting}[breaklines]
    You are an expert in classifying {problem} without data. Your role is to propose a single numeric value to divide the data into two groups based on the attribute {feature}. Don't analyze or reason too much, and deliver the numerical value directly. You must choose the value imagining that you have data to answer it. You must respect the restrictions given to you. If there is more than one restriction for the same attribute you must respect all restrictions.
\end{lstlisting}

Assistant message:

\begin{lstlisting}[breaklines]
Understood. I will reason and provide a numerical value based solely on my knowledge, ensuring the value of the attribute {feature} meets ALL the constraints: {branch_context}. I will provide a single numeric value to divide the data into two groups in a short response. If there is more than one restriction for the same attribute, all restrictions must be respected.
\end{lstlisting}

User message:

\begin{lstlisting}[breaklines]
What value of the attribute {feature} (that meets the constraints: {branch_context}) should I use to divide the data into two groups?\nChoose a single numeric value.
\end{lstlisting}

\subsubsection{Parse Numerical Features Splits}

System message:

\begin{lstlisting}[breaklines]
    You are an expert in extracting information from responses. Your task is to extract the numeric value suggested for dividing data into two groups based on a specific feature.
\end{lstlisting}

Assistant message:

\begin{lstlisting}[breaklines]
Understood. I will receive your response in the following format: Input:<response>. Then, I will extract and provide the numeric value after the prefix 'Output:'<numeric value>. For example: Input: 'For the numeric feature age, the best division is 15.5', I will respond 'Output: 15.5' without adding anything else. If no numeric value or grouping can be extracted, I will respond with the string 'Nothing'.
\end{lstlisting}

User message:

\begin{lstlisting}[breaklines]
For the numeric feature '{feature}', extract the numeric value indicating the cutoff point to divide the data into two groups. Input: {response_text}
\end{lstlisting}

\subsubsection{Get Categoric Features Splits}

System message:

\begin{lstlisting}[breaklines]
    You are an expert in {problem}. Your task is to use your knowledge to propose two disjoint groups of categories based on the categorical attribute {feature} with possible categories: {possible_values}, don't analyze or reason too much, delivers the groupings directly.. You have to try to make the groups formed as homogeneous as possible according to the {problem} variable that has as labels {target_labels}.
\end{lstlisting}

Assistant message:

\begin{lstlisting}[breaklines]
Understood. I will reason and provide two disjoint grpups of categories based solely on my knowledge, ensuring the groupings meet the constraints: {branch_context}. I will provide two disjoint groupings of categories to divide the data in a short response. I will try to to make the groups as homogeneous as possible.
\end{lstlisting}

User message:

\begin{lstlisting}[breaklines]
For the categorical attribute {feature} (that meets the constraints: {branch_context}) with possible values: {possible_values}, what two groups of categories should I use to divide the data into two groups as homogeneously as possible according to the {problem} variable with labels {target_labels}?\nChoose two disjoint groupings.
\end{lstlisting}

\subsubsection{Parse Categoric Features Splits}

System message:

\begin{lstlisting}[breaklines]
You are an expert in extracting information from responses. Your task is to extract the two disjoint groupings of categories suggested for dividing data into two groups based on a specific feature.
\end{lstlisting}

Assistant message:

\begin{lstlisting}[breaklines]
Understood. I will receive your response in the following format: Input:<response>. Then, I will extract and provide the two disjoint groupings after the prefix 'Output:'<grouping_1>;;<grouping_2>. For example: Input: 'For the categorical feature color, the best division is to create a group of colors [red, green] and another group of colors [blue]', I will respond 'Output: red, green;;blue' without adding anything else. If no groupings can be extracted, I will respond with the string 'Nothing'.
\end{lstlisting}

User message:

\begin{lstlisting}[breaklines]
For the categorical feature '{feature}' with possible values: {possible_values}, extract the two disjoint groupings of categories suggested for dividing the data into two groups. Input: {response_text}
\end{lstlisting}

\subsubsection{Get Probabilities Estimation}

System message:

\begin{lstlisting}[breaklines]
You are an expert in estimating the labels probabilities at the nodes of a decision tree. Your task is to provide a rough estimate of the probabilities that a given instance belongs to the possible target classes based on: the characteristics, probabilities of the previous node and the information from the new node split. Use only your general knowledge, since you do not have specific data. Remember that splits usually improve node purity. Remember also when there is little information do not give high probabilities (equal or higher than 0.9), unless you are very sure of them, because you may be overestimating.
\end{lstlisting}

Assistant message:

\begin{lstlisting}[breaklines]
Understood. Although it is difficult to provide exact probabilities for a particular individual, I will reason based on the information provided and my general knowledge to provide rough estimates of specific probabilities. I will provide probability values between 0 and 1 for each label, for each possible new division. In case there is little information, I will give low probabilities (< 0.9) unless I am very sure. On the other hand, I will not give high probabilities (>= 0.9) unless I am very sure, as I could be overestimating.
\end{lstlisting}

User message:

\begin{lstlisting}[breaklines]
Considering a classification problem of {problem}, estimate the probability that a {instance_type} has the target variable {target_feature} for each of the possible values: {classes}, based on the characteristics of the previous node: {previous_context} where the probability of the target variable is {previous_probabilities}, and the possible new divisions are: {division_1} or {division_2}.
\end{lstlisting}

\subsubsection{Parse Probabilities Estimation}

System message:

\begin{lstlisting}[breaklines]
You are an expert in extracting information from responses. Your task is to extract the label probability suggested for a classification problem to a specific label and new information. Do not provide anything other than the Output. Adhere to the format.
\end{lstlisting}

Assistant message:

\begin{lstlisting}[breaklines]
Understood. I will extract the label probability to a new information from the given response and provide it after the prefix 'Output:'<probability>. If no probability can be extracted, I will respond with 'Nothing'. For example: Input: 'The probability that a {instance_type} has the target variable with the value {class_1} is 0.8' Output: 0.8. Other example: Input: 'The probability that a {instance_type} has the target variable with the value {class_2} is 0.3' Output: 0.3.
\end{lstlisting}

User message:

\begin{lstlisting}[breaklines]
For the classification problem with the label {label} and the following new information {new_information}, extract the probability suggested for the problem. The given response is: {response_text}
\end{lstlisting}

\end{document}